\documentclass{article}

\usepackage{amsmath}
\usepackage{graphics}
\usepackage{makecell}
\usepackage{graphicx}
\usepackage{algorithm}
\usepackage{algpseudocode}
\usepackage{multirow}

\usepackage[final]{neurips_2022}

\usepackage[utf8]{inputenc} 
\usepackage[T1]{fontenc}    
\usepackage{hyperref}       
\usepackage{url}            
\usepackage{booktabs}       
\usepackage{amsfonts}       
\usepackage{nicefrac}       
\usepackage{microtype}      
\usepackage{xcolor}         

\title{Collective Knowledge Graph Completion with Mutual Knowledge Distillation}

\author{%
  Weihang Zhang \\
  Imperial College London\\
  \texttt{w.zhang21@imperial.ac.uk} \\
  \And
  Ovidiu Serban \\
  Imperial College London\\
  \texttt{o.serban@imperial.ac.uk} \\
  \AND
  Jiahao Sun \\
  Imperial College London\\
  Royal Bank of Canada\\
  \texttt{jiahao.sun@rbc.com} \\
  \And
  Yi-ke Guo \\
  Imperial College London\\
  \texttt{y.guo@imperial.ac.uk} \\
}

\begin{document}

\maketitle

\begin{abstract}
    Knowledge graph completion (KGC), the task of predicting missing information based on the existing relational data inside a knowledge graph (KG), has drawn significant attention in recent years. However, the predictive power of KGC methods is often limited by the completeness of the existing knowledge graphs from different sources and languages. In monolingual and multilingual settings, KGs are potentially complementary to each other. In this paper, we study the problem of multi-KG completion, where we focus on maximizing the collective knowledge from different KGs to alleviate the incompleteness of individual KGs. Specifically, we propose a novel method called CKGC-CKD that uses relation-aware graph convolutional network encoder models on both individual KGs and a large fused KG in which seed alignments between KGs are regarded as edges for message propagation. An additional mutual knowledge distillation mechanism is also employed to maximize the knowledge transfer between the models of ``global'' fused KG and the ``local'' individual KGs. Experimental results on multilingual datasets have shown that our method outperforms all state-of-the-art models in the KGC task.
\end{abstract}

\section{Introduction}

    Knowledge Graphs (KGs) are often viewed as large-scale semantic networks that store facts as triples in the form of \textit{(subject entity, relation, object entity)}. KGs have been widely adopted in many industrial applications because they capture the multi-relational nature between real-world entities well. However, real-world KGs are usually incomplete \citep{dong_knowledge_2014}. To tackle the incompleteness problem, there has been a surge of research interest in the KGC task in recent years. KGC and many other KG-based tasks are usually based on knowledge representation learning (KRL), in which entities and relations in a KG are encoded into low-dimensional vectors. With the recent advances in Graph Neural Network(GNN) \citep{scarselli_graph_2009}, many recently published methods like R-GCN \citep{schlichtkrull_modeling_2018} and CompGCN \citep{vashishth_composition-based_2020} all employed an encoder-decoder mechanism to tackle the KGC problem: variations of Graph Convolutional Networks (GCN) \citep{kipf_semi-supervised_2017} are used as encoders to generate embeddings for entities and relations in a KG, and traditional shallow KG embedding methods like TransE \citep{bordes_translating_2013} and DistMult \citep{yang_embedding_2015} are used as decoders for the KGC task. With the additional message propagation and aggregation mechanism of graph convolution operation in the encoding stage, these methods have shown more promising results on the KGC task than traditional knowledge graph embedding methods. However, not all the missing information can be inferred based on the information inside a KG. Even with a better encoding mechanism of GCNs, the expressiveness and quality of trained models can still be limited by the sparseness and incompleteness of the individual KG the model is trained on. At the same time, real-world entities are usually captured in more than one KG from either different sources or languages. The common entities in the disjoint real-world KGs can potentially serve as bridges to better connect them and transfer additional knowledge to one another to alleviate the sparseness problem faced by almost all real-world KGs. The common entities across different KGs are often known as \textit{seed alignments}, which usually originate from the manual annotation of human annotators. Because of the scale and size of KGs nowadays, seed alignments are usually relatively scarce. 
    
\begin{figure}
  \centering
  \includegraphics[width=0.5\textwidth]{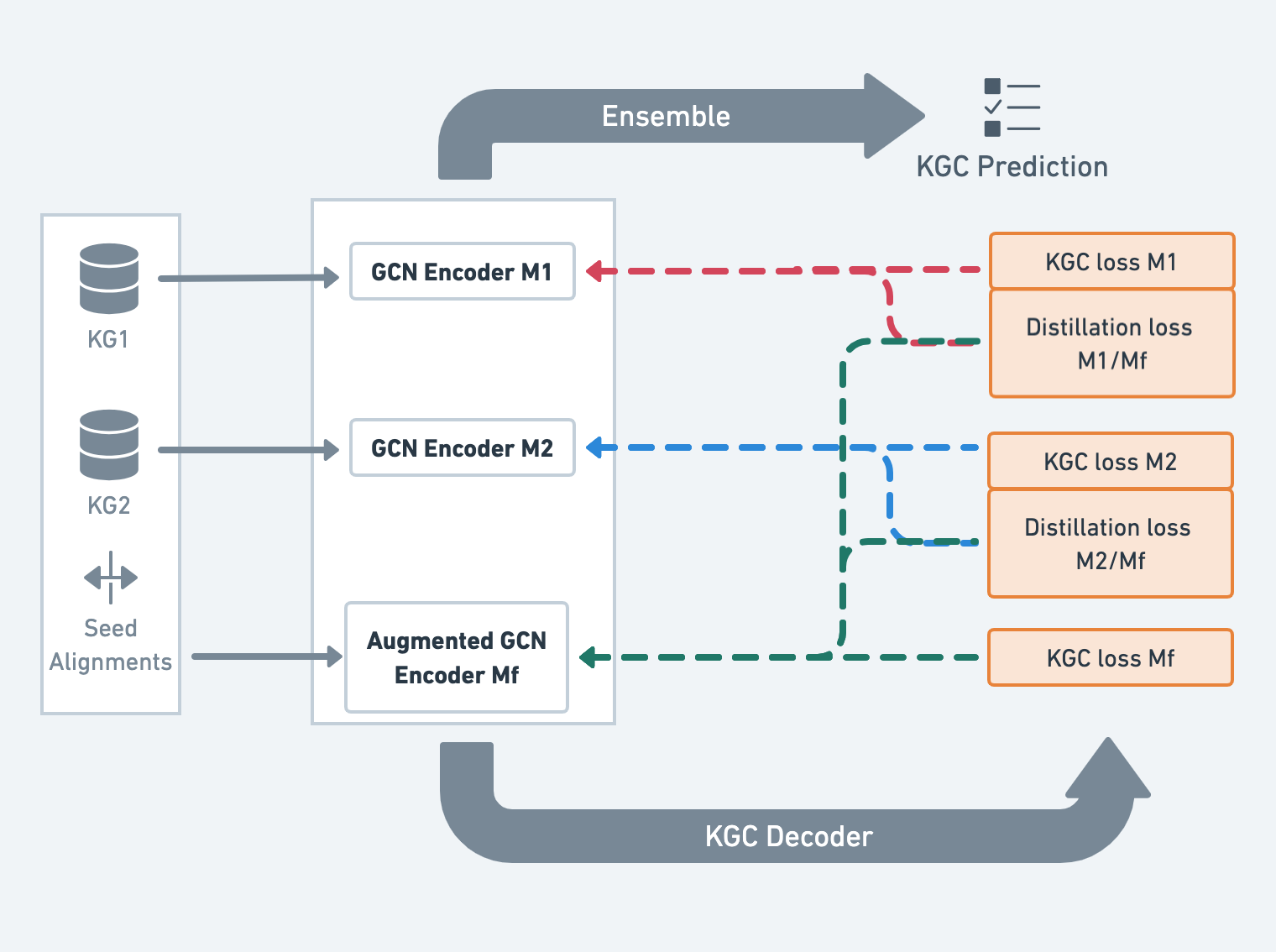}
  \caption{An illustrative figure of the proposed CKGC-CKD with 2 KGs.}
  \label{figure:illu}
\end{figure}

    In this paper, we focus on the multi-KG completion problem and aim to collectively utilize multiple KGs and seed alignments between them to maximize the KGC task performance on each individual KG. We propose a novel method that concurrently trains CompGCN-based \citep{vashishth_composition-based_2020} encoders on each individual KGs and a fused KG in which seed alignments are regarded as edges for connecting KGs together and augmented message propagation for ``knowledge transfer''. During the concurrent training, we also employ the mutual knowledge distillation mechanism. CompGCN-based encoders on individual KGs and the fused KG are trained to learn potentially complementary features from each other. The intuition behind the mutual knowledge distillation process is that the small encoders trained on individual KGs capture local semantic relationships while the large encoder trained on the large fused KG captures the global semantic relationships better because of the intra-KG message propagation. In the mutual knowledge distillation process, the small and large encoders take turns to become ``teachers'' in the knowledge distillation to encourage mutual knowledge transfer between them. Lastly, we use an ensemble to combine the predictions from the individual KG models and the fused KG model to produce the KGC predictions on the test set for each individual KG. Figure \ref{figure:illu} provides an illustrative figure of the overall structure of CKGC-CKD.
    
    The main contribution of this paper can be summarized as follows: 1) we propose a novel augmented CompGCN encoder to facilitate intra-KG knowledge transfer and tackle the multi-KG completion task; 2) we propose a novel mutual knowledge distillation mechanism to encourage collaborative knowledge transfer between the models trained on individual KGs and globally fused KG. Experimental results on popular multilingual datasets show that our proposed method outperforms all state-of-the-art models. Extensive ablation studies are conducted on monolingual and multilingual datasets to demonstrate the contribution of each component in the proposed method. 

\section{Related work}
\subsection{Knowledge graph embeddings}
    The research on knowledge graph embeddings has gained significant attention in recent years. This task aims to encode entities and relations of a KG into low-dimensional vectors. Traditional translation-based methods like TransE \citep{bordes_translating_2013}, TransH \citep{wang_knowledge_2014}, TransR \citep{lin_learning_2015}, as well as the semantic matching models like RESCAL \citep{nickel_three-way_2011} and DistMult \citep{yang_embedding_2015}, all achieved promising results on the KGC task. These methods all served as decoders on shallow embeddings to exploit the local structural information within triples and capture the semantics of relation triples. Other than the triple-based embedding models, another stream of recent works \citep{schlichtkrull_modeling_2018, vashishth_composition-based_2020, yu_knowledge_2021} all employed the graph structure to propagate information between adjacent entities and encode them into embeddings. Specifically, variants of the GCN model are used as encoders to embed entities and relations in the heterogeneous graphs into vectors. Traditional knowledge graph embedding methods like TransE are then used as decoders for various downstream tasks like KGC and node classification.
    
\subsection{KGC across multiple Knowledge graphs}
    Compared to KGC on a single KG, KGC across multiple KGs is a relatively under-explored area. \citet{wang_adversarial_2021} proposed ATransN, an adversarial embedding transfer network which aims to facilitate the knowledge transfer from a pre-trained embedding of a teacher KG to a student KG with a set of seed alignments. \citet{chen_multilingual_2020} was the first to propose multilingual KGC problem setting and tackled the problem by generating ensemble results on shared triples between KGs in different languages. On the same multilingual problem setting, \citet{singh_multilingual_2021} proposed AlignKGC, which employs a multi-task strategy to jointly trains KGC, entity alignment and relation alignment tasks. Most recently, \citet{huang_multilingual_2022} proposed SS-AGA, which models seed alignments as edges to fuse multiple knowledge graphs while using a generator model to dynamically capture more potential alignments between entities and iteratively add more edges to the graph. Additionally, \citet{sourty_knowledge_2020} proposed KD-MKB, which assumes the existence of both shared relations and shared entities across individual KGs, and therefore tackles multi-KG completion tasks from a knowledge distillation perspective.

\section{Methods}
\subsection{Preliminaries}
     The framework of a multi-KG completion task involves two or more KGs. Without loss of generality, we assume there are a total of $m$ KGs in the problem setting. We formalize the $i$-th heterogeneous KG in the task as $KG_i = \{E_i, R_i, T_i\}$, where ${E_i, R_i, T_i}$ respectively represent the entity set, relation set, and fact triple set of $KG_i$. A set of seed alignments between KGs, known before training, is denoted by $S_{KG_i, KG_j} = \{(e_i, \sim, e_j): (e_i, e_j) \in E_i \times E_j\}$, where $\sim$ denotes the equivalence relation. The complete set of seed alignments can then be denoted by $S_{align} = \cup_{i=1}^{m}\cup_{j=i+1}^{m} S_{KG_i, KG_j}$. We can then formalize the large fused KG connected by seed alignments as $KG_f =\{E_f, R_f, T_f | E_f = \cup_{i=1}^{m}E_i, R_f = \cup_{i=1}^{m}R_i, T_f = (\cup_{i=1}^{m}T_i) \cup S_{align} \}$. Let $M_{i}$ and $M_{f}$ denote the encoder models for the $i$-th individual KG and the fused KG, respectively.

\subsection{Augmented CompGCN Link Prediction}
    We use CompGCN \citep{vashishth_composition-based_2020} as our encoders for the knowledge graph embeddings. In the method, CompGCN encoders with non-parametric composition operations are trained on each individual KGs and the fused KG concurrently. In standard CompGCN, the update equation of CompGCN node embeddings can be written as: \begin{align}\label{eq:compupdate}
    h^t_v = f(\Sigma_{(u, r) \in N(v)} Me(u, r))
    \end{align}
    \begin{align}\label{eq:compmessage}
    Me(u, r) = W_{\lambda(r)} \phi(h^{t-1}_u, h^{t-1}_r)
    \end{align}
    , where $h^t_v$ denotes the updated embedding for node $v$ at $t$-th layer, $N(v)$ denotes the set of neighboring entities and relations of node $v$, $h^{t-1}_u$ and $h^{t-1}_r$ denotes the embeddings for node $u$ and relation $r$ at ($t$-$1$)-th layer respectively, $\phi$ denotes the non-parametric composition operation and $W_{\lambda(r)}$ denotes the direction specific transformation matrix where $\lambda$ denotes the direction of relation. In our method, the vanilla CompGCN encoder is used without modification on individual KGs, while we decide to use an augmented CompGCN encoder for better knowledge transfer on the fused $KG_f$. Specifically, although seed alignments are viewed as regular relations in the fused KG, we remove the composition operator for message propagation between the seed alignments and instead use the standard non-relation-specific message passing between KGs. The augmented message function in the fused KG can then be written as: \begin{align}\label{eq:compaug}
    Me(u, r) = 
    \begin{cases} 
        W_{align} h^{t-1}_u \qquad \quad \text{if} (u, \sim, v) \in S_{align} \\
        W_{\lambda(r)} \phi(h^{t-1}_u, h^{t-1}_r) \quad  \text{otherwise}
    \end{cases}
    \end{align}, where $W_{align}$ denotes the transformation matrix specifically for intra-KG message propagation. The composition operation is removed because we view the cross-KG equivalence as a bi-directional relationship different from the triples inside KGs. Additionally, many existing methods \citep{wang_adversarial_2021, singh_multilingual_2021} use a loss regularization to ensure the equivalent entities in each KG have similar embeddings with or without transformation. However, instead of imposing the regularization directly on the training loss term, we impose a softer regularization in the message passing augmentation, where the contextualized node embeddings of entities in each knowledge graph are passed to their counterparts in other KGs during encoding. As a result, the contextualized embedding of entities in each KG can be shared across the KGs by the augmented message propagation in the encoding phase and optimized during the training of the KGC task on fact triples. 
    
    The encoded entity and relation embeddings are then passed to the decoder, which performs the link prediction task on triples in KG, and computes the knowledge representation loss. The margin-based knowledge representation loss can be written as: \begin{align}\label{eq:kr}
    L_T = \Sigma_{t_i \in T_i, t_i'\in T_i'} f(t_i) - f(t_i') + \gamma
    \end{align}
    , where $T_i'$ denotes the negative samples created from corrupting head or tail entity in triple $t_i$; $f(t_i)$ denotes the scoring function for TransE or similar knowledge embedding model; and $\gamma$ denotes the margin, which is a hyperparameter describing the ideal distance between positive triples and negative triples.
    
\subsection{Mutual Knowledge Distillation}
    We employ the mutual knowledge distillation mechanism between each model on individual KGs $M_i$ and the model on the fused KG $M_f$. At each training step, each $M_i$ pairs with $M_f$ to conduct mutual knowledge distillation, where $M_i$ and $M_f$ learn simultaneously from each other via a mimicry loss that measures the difference between the posterior predictions of each other on the KGC task on triples $T_i$ in the corresponding $KG_i$. Three different KGC tasks are used for mutual knowledge distillation: for a triple $t = (s, r, o)$, the task is to predict the missing component given the other two in the triple, i.e., head prediction, tail prediction and relation prediction. The distillation loss can be written as: \begin{align}\label{eq:distill}
    L_{D}^i = \sum_{t_i \in T_i} \sum_{\beta \in Task} 
        d_{KL} (t_i, \beta)
    \end{align}
    \begin{align}
        d_{KL} (t_i, \beta) = D_{KL}(P_{i}^{\beta}(t_i), P_{f}^{\beta}(t_i))
    \end{align}
    , where $Task$ denotes three KGC tasks, $D_{KL}$ denotes the Kullback Leibler (KL) Divergence, and $P$ denotes the categorical distribution predicted by the knowledge graph embedding scoring function on task $\beta$. As an example, for tail prediction of triple $t_i = (s_i, r_i, o_i)$, the categorical distribution can be written as softmax of tail prediction across all candidates: \begin{align}
        P_{i}(t_i) = \frac{exp(f(M_i(s_i), M_i(r_i), M_i(o_i)))}{\sum_{o_j \in E_i} exp(f(M_i(s_i), M_i(r_i), M_i(o_j)))}
    \end{align}, where $M_i()$ denotes the embedding look-up operation for entities and relations from encoder model $M_i$. In practice, predicting across all candidates $E_i$ and comparing the categorical distribution across all entities can be inefficient due to the size of KG. Therefore, we employ the top-k sampling technique used in the work of \citet{sourty_knowledge_2020} by using the ``teacher'' model to select the top-k most confident candidates for the categorical distribution comparison in the mutual distillation. 
    
    Additionally, we enabled the mutual distillation process to be performance-aware as the performance of individual model $M_i$ and the fused model $M_f$ could differ by some margin. A recent work \citep{xie_performance-aware_2022} on knowledge distillation pointed out that a worse-performing model could potentially generate negative knowledge transfer and lead to collective failure. Therefore, we adopted a softer restriction than \citet{xie_performance-aware_2022} and designed a hyperparameter $\theta$ such that the worse-performing model is allowed to become a teacher and generate soft training target only if its mean reciprocal rank metric on the validation set is less $\theta$ away from the better-performing model.
    
\subsection{Training}
    The overall loss term combines the knowledge representation and mutual knowledge distillation loss: \begin{align}
    L = L_T + \alpha L_D 
    \end{align}, where $\alpha$ is a hyperparameter controlling the contribution of mutual distillation loss to the overall loss term. The models $M_i$ and $M_f$ are trained concurrently on KGC tasks while learning from the best-performing model of each other via the mutual distillation process. In practice, the training process is separated into two stages for better convergence and faster training. In the first stage, individual and fused models are trained independently with only knowledge representation loss, while in the second stage, knowledge distillation losses are introduced so that models can mutually learn from each other. 

\subsection{Ensemble Prediction}
    In the end, the output for KGC tasks is generated by combining predictions from models $M_i$ and $M_f$ using an ensemble. Concretely, the for triple $t_i \in T_i$, the final scoring function becomes: $F(t_i) = f(M_i(t_i)) + f(M_f(t_i))$. The ensemble scores are then used for further ranking and evaluation.

\begin{table}[]
\caption{Results on DBP-5L dataset.}
\label{table:multiresults}
\centering
\resizebox{\columnwidth}{!}{
\begin{tabular}{l|c|c|c|c|c}
\hline
         & 
         \multicolumn{1}{c|}{\textbf{EL}} &
         \multicolumn{1}{c|}{\textbf{JA}} &
         \multicolumn{1}{c|}{\textbf{ES}} &
         \multicolumn{1}{c|}{\textbf{FR}} & 
         \multicolumn{1}{c}{\textbf{EN}} \\ \hline
         & \multicolumn{1}{c|}{H@1/H@10/MRR} &
         \multicolumn{1}{c|}{H@1/H@10/MRR} &
         \multicolumn{1}{c|}{H@1/H@10/MRR} &
         \multicolumn{1}{c|}{H@1/H@10/MRR} &
         \multicolumn{1}{c}{H@1/H@10/MRR} \\ \hline
\textbf{KenS}     &  28.1 / 56.9 / -    & 32.1 / 65.3 / -    & 23.6 / 60.1 / -    & 25.5 / 62.9 / -    & 15.1 / 39.8 / - \\ \hline
\textbf{CG-MuA}   &  21.5 / 44.8 / 32.8 & 27.3 / 61.1 / 40.1 & 22.3 / 55.4 / 34.3 & 24.2 / 57.1 / 36.1 & 13.1 / 33.5 / 22.2 \\ \hline
\textbf{AlignKGC} &  27.6 / 56.3 / 33.8 & 31.6 / 64.3 / 41.6 & 24.2 / 60.9 / 35.1 & 24.1 / 62.3 / 37.4 & 15.5 / 39.2 / 22.3 \\ \hline
\textbf{SS-AGA}   &  30.8 / 58.6 / 35.3 & 34.6 / 66.9 / 42.9 & 25.5 / 61.9 / 36.6 & 27.1 / 65.5 / 38.4 & 16.3 / 41.3 / 23.1 \\ \Xhline{1.5pt}
\textbf{KGC-I} & 28.9 / 66.8 / 41.6 & 30.3 / 61.7 / 41.4 & 24.8 / 61.2 / 37.5 & 25.8 / 64.1 / 39.1 & 20.5 / 58.6 / 33.5 \\ \hline
\textbf{KGC-A} & 43.4 / \textbf{89.2} / 60.1 & 42.7 / 83.6 / 57.2 & 32.8 / 77.2 / 48.4 & 35.1 / 81.4 / 51.3 & 25.1 / \textbf{67.2} / 39.6\\ \hline
\textbf{CKGC-CKD} &  \textbf{49.2} / 88.5 / \textbf{63.9} & \textbf{48.4} / \textbf{84.6} / \textbf{60.9} & \textbf{37.5} / \textbf{77.6} / \textbf{51.8} & \textbf{40.9} / \textbf{83.0} / \textbf{55.8} & \textbf{28.8} / 67.1 / \textbf{42.1} \\ \hline
\end{tabular}
}
\end{table}

\begin{table}[]
\caption{Results on E-PKG dataset.}
\label{table:multiresults2}
\centering
\resizebox{\columnwidth}{!}{
\begin{tabular}{l|c|c|c|c|c|c}
\hline
         & 
         \multicolumn{1}{c|}{\textbf{EN}} &
         \multicolumn{1}{c|}{\textbf{DE}} &
         \multicolumn{1}{c|}{\textbf{FR}} &
         \multicolumn{1}{c|}{\textbf{JA}} &
         \multicolumn{1}{c|}{\textbf{ES}} & 
         \multicolumn{1}{c}{\textbf{IT}} \\ \hline
         & \multicolumn{1}{c|}{H@1/H@10/MRR} &
         \multicolumn{1}{c|}{H@1/H@10/MRR} &
         \multicolumn{1}{c|}{H@1/H@10/MRR} &
         \multicolumn{1}{c|}{H@1/H@10/MRR} &
         \multicolumn{1}{c|}{H@1/H@10/MRR} &
         \multicolumn{1}{c}{H@1/H@10/MRR} \\ \hline
\textbf{KenS}     &  26.2 / 69.5 / - & 24.3 / 65.8 / -   & 25.4 / 68.2 / -   & 33.5 / 73.6 / -    & 21.3 / 59.5 / -  & 25.1 / 64.6 / -     \\ \hline
\textbf{CG-MuA}   &  24.8 / 67.9 / 40.2 & 22.9 / 64.9 / 38.7 & 23.0 / 67.5 / 39.1 & 30.4 / 72.9 / 45.9 & 19.2 / 58.8 / 33.8 & 23.9 / 63.8 / 37.6 \\ \hline
\textbf{AlignKGC} &  25.6 / 68.3 / 40.5 & 22.1 / 65.1 / 38.5 & 22.8 / 67.2 / 38.8 & 31.2 / 72.3 / 46.2 & 19.4 / 59.1 / 34.2 & 24.2 / 63.4 / 37.3 \\ \hline
\textbf{SS-AGA}   &  26.7 / 69.8 / 41.5 & 24.6 / 66.3 / 39.4 & 25.9 / 68.7 / 40.2 & 33.9 / 74.1 / 48.3 & 21.0 / 60.1 / 36.3 & 24.9 / 63.8 / 38.4 \\ \Xhline{1.5pt}
\textbf{KGC-I}    & 44.5 / 81.7 / 58.6 & 36.3 / 75.4 / 50.4 & 36.5 / 77.4 / 51.2 & 44.5 / 83.5 / 59.3 & 30.2 / 67.3 / 43.6 & 37.4 / 75.1 / 51.6 \\ \hline
\textbf{KGC-A}    & 44.3 / 82.8 / 58.8 & 38.5 / 78.7 / 53.1 & 37.1 / 79.9 / 52.5 & 46.2 / 84.7 / 60.8 & 36.1 / 73.8 / 49.6 & 40.1 / 78.4 / 54.3 \\ \hline
\textbf{CKGC-CKD} & \textbf{46.6} / \textbf{84.1} / \textbf{60.7} & \textbf{40.0} / \textbf{79.8} / \textbf{54.4} & \textbf{39.0} / \textbf{81.0} / \textbf{54.2} & \textbf{47.5} / \textbf{85.5} / \textbf{62.1} & \textbf{36.9} / \textbf{75.3} / \textbf{50.5} & \textbf{41.6} / \textbf{80.0} / \textbf{55.8} \\ \hline
\end{tabular}
}
\end{table}

\section{Experiments}
\label{sec:expr}
\subsection{Basic settings}
    All of our experiments are conducted on a Linux server running Ubuntu 20.04 with 64GB memory and a Nvidia A100 GPU.
    
    We perform experiments and compare the performance of the proposed CKGC-CKD method with the state-of-the-art models on the existing multilingual dataset \textbf{DBP-5L} \citep{chen_multilingual_2020} and \textbf{E-PKG} \citep{huang_multilingual_2022}. \textbf{DBP-5L} is a dataset sampled from DBpedia \citep{auer_dbpedia_2007}; it contains five KGs from different languages: English (EN), French (FR), Spanish (ES), Japanese (JA) and Greek (EL). E-PKG is an industrial multilingual E-commerce product KG dataset, which describes phone-related products across six different languages: English (EN), German (DE), French(FR), Japanese (JA), Spanish (ES) and Italian (IT). In this work, we follow the evaluation scheme of previous works \citep{chen_multilingual_2020, singh_multilingual_2021, huang_multilingual_2022}: for a test triple $(h, r, t)$, use the embedding model to rank all possible answers to tail prediction query $(h, r, ?)$; and apply the MRR(mean reciprocal ranks), Hit@1 and Hit@10 metrics under filtered settings \citep{wang_knowledge_2014, yang_embedding_2015} to evaluate the performance of the models. The reported CKGC-CKD model uses a 1-layer encoder, with TransE as a knowledge embedding decoder and an embedding dimension of 100. However, CKGC-CKD can be easily extended to use other knowledge-embedding decoders. Deeper encoder can also be explored if more hardware resources become available.

\begin{table}[]
\caption{Ablation study results on DBP-5L and D-W-15K-LP.}
\label{table:ablation}
\centering
\scalebox{0.9}{
\begin{tabular}{l|l|cccccccc}
\cline{1-9}
 & & \multicolumn{5}{c|}{\textbf{DBP-5L}} & \multicolumn{2}{c}{\textbf{D-W-15K-LP}} &  \\ \cline{1-9}
 & \multicolumn{1}{l|}{\textbf{Metric}}                                                   & \multicolumn{1}{c|}{\textbf{EL}}                                                         & \multicolumn{1}{c|}{\textbf{JA}}                                                         & \multicolumn{1}{c|}{\textbf{ES}}                                                         & \multicolumn{1}{c|}{\textbf{FR}}                                                         & \multicolumn{1}{c|}{\textbf{EN}}                                                         & \multicolumn{1}{c|}{\textbf{DBpedia}}                                                    & \multicolumn{1}{c}{\textbf{Wikidata}} &  \\ \cline{1-9}
\textbf{KGC-I}    & \multicolumn{1}{l|}{\begin{tabular}[c]{@{}l@{}}H@1\\ H@10\\ MRR\end{tabular}} & \multicolumn{1}{c|}{\begin{tabular}[c]{@{}c@{}}22.0\\ 49.1\\ 31.3\end{tabular}} & \multicolumn{1}{c|}{\begin{tabular}[c]{@{}c@{}}23.1\\ 44.7\\ 30.7\end{tabular}} & \multicolumn{1}{c|}{\begin{tabular}[c]{@{}c@{}}19.0\\ 44.1\\ 27.9\end{tabular}} & \multicolumn{1}{c|}{\begin{tabular}[c]{@{}c@{}}21.1\\ 45.3\\ 29.5\end{tabular}} & \multicolumn{1}{c|}{\begin{tabular}[c]{@{}c@{}}17.2\\ 45.5\\ 26.9\end{tabular}} & \multicolumn{1}{c|}{\begin{tabular}[c]{@{}l@{}}29.9\\ 54.2\\ 38.4\end{tabular}} & \multicolumn{1}{c}{\begin{tabular}[c]{@{}l@{}}25.5\\ 49.2\\ 34.2\end{tabular}} &  \\ \cline{1-9}
\textbf{KGC-C}    & \multicolumn{1}{l|}{\begin{tabular}[c]{@{}l@{}}H@1\\ H@10\\ MRR\end{tabular}} & \multicolumn{1}{c|}{\begin{tabular}[c]{@{}c@{}}36.4\\ 69.5\\ 47.9\end{tabular}} & \multicolumn{1}{c|}{\begin{tabular}[c]{@{}c@{}}32.7\\ 63.6\\ 43.6\end{tabular}} & \multicolumn{1}{c|}{\begin{tabular}[c]{@{}c@{}}27.0\\ 57.1\\ 37.6\end{tabular}} & \multicolumn{1}{c|}{\begin{tabular}[c]{@{}c@{}}28.7\\ 59.7\\ 39.5\end{tabular}} & \multicolumn{1}{c|}{\begin{tabular}[c]{@{}c@{}}20.2\\ 50.8\\ 30.7\end{tabular}} & \multicolumn{1}{c|}{\begin{tabular}[c]{@{}l@{}}29.9\\ 55.3\\ 38.8\end{tabular}} & \multicolumn{1}{c}{\begin{tabular}[c]{@{}l@{}}26.8\\ 50.4\\ 35.4\end{tabular}} &  \\ \cline{1-9}
\textbf{KGC-A}    & \multicolumn{1}{l|}{\begin{tabular}[c]{@{}l@{}}H@1\\ H@10\\ MRR\end{tabular}} & \multicolumn{1}{c|}{\begin{tabular}[c]{@{}c@{}}37.9\\ 71.5\\ 49.9\end{tabular}} & \multicolumn{1}{c|}{\begin{tabular}[c]{@{}c@{}}35.5\\ 67.0\\ 46.5\end{tabular}} & \multicolumn{1}{c|}{\begin{tabular}[c]{@{}c@{}}28.0\\ 59.5\\ 39.1\end{tabular}} & \multicolumn{1}{c|}{\begin{tabular}[c]{@{}c@{}}29.4\\ 62.2\\ 40.9\end{tabular}} & \multicolumn{1}{c|}{\begin{tabular}[c]{@{}c@{}}21.0\\ 53.8\\ 32.3\end{tabular}} & \multicolumn{1}{c|}{\begin{tabular}[c]{@{}l@{}}30.7\\ 55.6\\ 39.3\end{tabular}} & \multicolumn{1}{c}{\begin{tabular}[c]{@{}l@{}}27.5\\ 50.8\\ 35.8\end{tabular}} &  \\ \cline{1-9}
\textbf{KGC-I-D}  & \multicolumn{1}{l|}{\begin{tabular}[c]{@{}l@{}}H@1\\ H@10\\ MRR\end{tabular}} & \multicolumn{1}{c|}{\begin{tabular}[c]{@{}c@{}}37.8\\ 66.0\\ 48.1\end{tabular}} & \multicolumn{1}{c|}{\begin{tabular}[c]{@{}c@{}}35.4\\ 62.4\\ 44.9\end{tabular}} & \multicolumn{1}{c|}{\begin{tabular}[c]{@{}c@{}}28.6\\ 55.4\\ 38.0\end{tabular}} & \multicolumn{1}{c|}{\begin{tabular}[c]{@{}c@{}}30.5\\ 58.8\\ 40.6\end{tabular}} & \multicolumn{1}{c|}{\begin{tabular}[c]{@{}c@{}}22.3\\ 50.3\\ 31.8\end{tabular}} & \multicolumn{1}{c|}{\begin{tabular}[c]{@{}l@{}}31.2\\ 54.9\\ 39.4\end{tabular}} & \multicolumn{1}{c}{\begin{tabular}[c]{@{}l@{}}29.2\\ 49.8\\ 36.5\end{tabular}} &  \\ \cline{1-9}
\textbf{KGC-A-D}  & \multicolumn{1}{l|}{\begin{tabular}[c]{@{}l@{}}H@1\\ H@10\\ MRR\end{tabular}} & \multicolumn{1}{c|}{\begin{tabular}[c]{@{}c@{}}40.2\\ \textbf{71.8}\\ 51.3\end{tabular}} & \multicolumn{1}{c|}{\begin{tabular}[c]{@{}c@{}}37.8\\ 66.8\\ 47.9\end{tabular}} & \multicolumn{1}{c|}{\begin{tabular}[c]{@{}c@{}}30.2\\ \textbf{59.5}\\ 40.6\end{tabular}} & \multicolumn{1}{c|}{\begin{tabular}[c]{@{}c@{}}31.9\\ 63.1\\ 42.8\end{tabular}} & \multicolumn{1}{c|}{\begin{tabular}[c]{@{}c@{}}23.0\\ 52.7\\ 33.3\end{tabular}} & \multicolumn{1}{c|}{\begin{tabular}[c]{@{}l@{}}\textbf{31.7}\\ 55.7 \\ 40.0\end{tabular}} & 
\multicolumn{1}{c}{\begin{tabular}[c]{@{}l@{}}29.2\\ 50.7\\ 36.9\end{tabular}} &  \\ \cline{1-9}
\textbf{CKGC-CKD} & \multicolumn{1}{l|}{\begin{tabular}[c]{@{}l@{}}H@1\\ H@10\\ MRR\end{tabular}} & \multicolumn{1}{c|}{\begin{tabular}[c]{@{}c@{}}\textbf{41.4}\\ 70.5\\ \textbf{52.1}\end{tabular}} & \multicolumn{1}{c|}{\begin{tabular}[c]{@{}c@{}}\textbf{38.7}\\ \textbf{67.2}\\ \textbf{48.6}\end{tabular}} & \multicolumn{1}{c|}{\begin{tabular}[c]{@{}c@{}}\textbf{31.3}\\ \textbf{59.5}\\ \textbf{41.2}\end{tabular}} & \multicolumn{1}{c|}{\begin{tabular}[c]{@{}c@{}}\textbf{32.7}\\ \textbf{63.4}\\ \textbf{43.6}\end{tabular}} & \multicolumn{1}{c|}{\begin{tabular}[c]{@{}c@{}}\textbf{23.2}\\ \textbf{53.4}\\ \textbf{33.6}\end{tabular}} & \multicolumn{1}{c|}{\begin{tabular}[c]{@{}l@{}}\textbf{31.7}\\ \textbf{55.8}\\ \textbf{40.1}\end{tabular}} & 
\multicolumn{1}{c}{\begin{tabular}[c]{@{}l@{}}\textbf{29.3}\\ \textbf{50.8} \\ \textbf{37.0}\end{tabular}} &  \\ \cline{1-9}
\end{tabular}
}
\end{table}

\subsection{Results}
    In table \ref{table:multiresults} and \ref{table:multiresults2}, we present the experiment results on the \textbf{DBP-5L} dataset \citep{chen_multilingual_2020} and \textbf{E-PKG} dataset \citep{huang_multilingual_2022} respectively \footnote{We directly report the benchmarking results from the work of \citet{huang_multilingual_2022} for the first four rows in the table. However, we discovered that the filtered setting used in the benchmark was slightly different from the traditional filtered setting: \citet{huang_multilingual_2022} assumes the candidate space during testing excludes all positive triples from the training set. At the same time, the traditional filtered setting also excludes validation and test set. For fairness of comparison, all results we report in the table adopted the filtered setting by \citet{huang_multilingual_2022}.}. In the tables, performances of two extra baseline models are reported: KGC-I refers to the standard CompGCN encoder model trained on individual KG, and KGC-A refers to the augmented message propagation encoder trained on the fused KG. It can be observed that the proposed CKGC-CKD method outperforms all baseline and state-of-the-art models on both datasets. Compared to the previous cross-lingual models, the individually trained KGC-I model on each language can already achieve similar performance in most languages, which verifies the effectiveness of the CompGCN encoder. The KGC-A model trained on the fused KG provided a large margin over the KGC-I and the previous models. This implies that the inclusion of multiple KGs truly helps the KGC task of each other and also verifies the benefit of the augmented cross-KG message propagation even without explicit data leakage (direct swapping of relations between aligned entities). In the end, with mutual knowledge distillation between KGC-I and KGC-A enabled, the CKGC-CKD model uses the ensemble predictions from both distilled models. This achieves the best performances in the table across almost all of the metrics. Overall, compared to the KGC-I model trained on monolingual KG, CKGC-CKD achieves a significant gain on the lower-resource languages like Greek(EL) in DBL-5L and Japanese(JA) in E-PKG, which verifies the common belief that lower-resource KGs would typically benefit from the multi-KG learning setting. Moreover, we also observed performance improvement on relatively rich languages like French(FR) and English(EN) also in the experiments.

\section{Ablation study}

\subsection{Contribution of each component}
    In table \ref{table:ablation}, we report the results of our ablation studies to analyze how each of the components in the proposed method affects the results. We report the ablation study results on the multilingual DBP-5L dataset and a monolingual self-generated D-W-15K-LP dataset. D-W-15K-LP is a dataset generated from the entity alignment benchmarking datasets D-W-15K \citep{sun_benchmarking_2020}. To mimic a more real-life setting, we employed the sampling strategy proposed in the work of \citet{sun_knowing_2021} to create dangling entities (entities without alignment across KGs) in the KGs. In the sampling process, triples containing removed entities are excluded by removing part of the alignments from KGs. This results in a more sparse dataset with dangling entities in each individual KG. In addition to the KGC-I and the KGC-A models reported in the section \ref{sec:expr}, we additionally report the performance of several ablation models: KGC-C refers to the ablation model trained on fused KG without augmented message propagation, KGC-I-D and KGC-A-D respectively represent the ablation models with mutual distillation enabled for KGC-I and KGC-A. Therefore, the reported CKGC-CKD is the ensemble results of KGC-A-D and KGC-I-D. For a more complete and universal comparison, in the ablation study, we use the traditional ``link prediction'' task that includes both head prediction and tail prediction with the traditional filtered setting used in the works of \citet{wang_knowledge_2014} and \citet{yang_embedding_2015}. 
    
    On both datasets, we can observe a clear margin that the KGC-A model created over the KGC-C model, which verifies the effectiveness of augmented message propagation. Additionally, on both datasets, the distillation-enabled KGC-I-D and KGC-A-D models have shown superior performance in almost all metrics over the KGC-I and KGC-A models, respectively. This has shown that the mutual knowledge distillation process benefits individual and fused models. Lastly, CKGC-CKD achieves the best performances in most metrics, which verifies the gains provided by the ensemble technique. An interesting observation is that even after the mutual knowledge distillation, the KGC-I-D models still perform slightly worse than the fused model KGC-A-D, and the difference in performance also varies across different KG. One possible reason behind this observation is that we used a constant $\alpha$ for all KGs in one dataset to control the trade-off between knowledge distillation loss and knowledge representation loss. Limited by the hardware resources, we did not explore possibilities of assigning different $\alpha$ for each KG and decided to leave that for future work that possibly explores a fine-tuning stage of the model to better reconcile the difference and imbalance of resources in each KG.

\begin{table}[]
\caption{Results of CKGC-CKD models w/ and w/o extra triples introduced by alignment-informed meta-paths, metrics reported on link prediction task under standard filtered settings.}
\label{table:meta}
\centering
\resizebox{\columnwidth}{!}{
\begin{tabular}{l|c|c|c|c|c}
\hline
         \multicolumn{1}{c|}{\multirow{2}{*}{\textbf{DBP-5L}}} & 
         \multicolumn{1}{c|}{\textbf{EL}} &
         \multicolumn{1}{c|}{\textbf{JA}} &
         \multicolumn{1}{c|}{\textbf{ES}} &
         \multicolumn{1}{c|}{\textbf{FR}} & 
         \multicolumn{1}{c}{\textbf{EN}} \\ \cline{2-6}
         & \multicolumn{1}{c|}{H@1/H@10/MRR} &
         \multicolumn{1}{c|}{H@1/H@10/MRR} &
         \multicolumn{1}{c|}{H@1/H@10/MRR} &
         \multicolumn{1}{c|}{H@1/H@10/MRR} &
         \multicolumn{1}{c}{H@1/H@10/MRR} \\ \hline
\textbf{w/ extra triples}     &  \textbf{49.8} / \textbf{80.8} / \textbf{61.0}   & \textbf{44.6} / \textbf{74.8} / \textbf{55.51} &    \textbf{39.3} / \textbf{69.0} / \textbf{50.0} &   \textbf{39.9} / \textbf{70.9} / \textbf{51.7}  &   \textbf{28.3} / \textbf{59.4} / \textbf{39.2} \\ \hline
\textbf{w/o extra triples}   &  41.4 / 70.5 / 52.1   & 38.7 / 67.2 / 48.6    & 31.3 / 59.5 / 41.2    & 32.7 /63.4 / 43.6 & 23.2 / 53.4 / 33.6 \\ \hline
\end{tabular}
}

\resizebox{\columnwidth}{!}{
\begin{tabular}{l|c|c|c|c|c|c}
\hline
         \multicolumn{1}{c|}{\multirow{2}{*}{\textbf{E-PKG}}} & 
         \multicolumn{1}{c|}{\textbf{EN}} &
         \multicolumn{1}{c|}{\textbf{DE}} &
         \multicolumn{1}{c|}{\textbf{FR}} &
         \multicolumn{1}{c|}{\textbf{JA}} &
         \multicolumn{1}{c|}{\textbf{ES}} & 
         \multicolumn{1}{c}{\textbf{IT}} \\ \cline{2-7}
         & \multicolumn{1}{c|}{H@1/H@10/MRR} &
         \multicolumn{1}{c|}{H@1/H@10/MRR} &
         \multicolumn{1}{c|}{H@1/H@10/MRR} &
         \multicolumn{1}{c|}{H@1/H@10/MRR} &
         \multicolumn{1}{c|}{H@1/H@10/MRR} &
         \multicolumn{1}{c}{H@1/H@10/MRR} \\ \hline
\textbf{w/ extra triples}     &  42.4 / 61.8 / 49.3 & 21.8 / 42.3 / 29.4 & 24.9 / 43.3 / 32.0 & 22.9 / 42.9 / 30.5 & 22.9 / 44.5 / 31.1 & 28.4 / 46.4 / 35.4     \\ \hline
\textbf{w/o extra triples}   & \textbf{46.0} / \textbf{66.8} / \textbf{53.6} & \textbf{23.6} / \textbf{46.0} / \textbf{31.9} & \textbf{26.1} / \textbf{46.4} / \textbf{33.9} & \textbf{24.5} / \textbf{45.5} / \textbf{32.5} & \textbf{24.1} / \textbf{46.1} / \textbf{32.5} & \textbf{29.3} / \textbf{48.5} / \textbf{36.7} \\ \hline
\end{tabular}
}
\end{table}

\subsection{Discussion on the alignment-informed meta-path}
    With multilingual datasets sharing the same relation schema, there are obvious mechanisms to introduce additional triples information via an alignment-informed meta-path. For example, a meta-path of ($E_a \xrightarrow{SAMEAS} E_b \xrightarrow{R_a} E_c \xrightarrow{SAMEAS} E_d$) implies a new triple $(E_a \xrightarrow{R_a} E_d)$, this is usually referred as parameter-swapping \citep{sun_benchmarking_2020}. When the dataset involves more than two KGs, additional alignments can potentially also be inferred via meta-path: $(E_a \xrightarrow{SAMEAS} E_b \xrightarrow{SAMEAS} E_c)$ implies $(E_a \xrightarrow{SAMEAS} E_c)$. Additional triples information can be generated or shared using the two alignment-informed meta-paths above. However, this relies heavily on the quality annotation of seed alignments between the KGs, which is not usually guaranteed, especially in scenarios involving multiple KGs. In table \ref{table:meta}, we report the link prediction performances (incl. both head prediction and tail prediction) of our method with and without alignment-informed meta-path included in training on the two multilingual datasets, in which we have observed fairly contradictory phenomenons: DBP5L dataset benefits massively from the additional triples from alignment-informed meta-path, while on the E-PKG dataset we have observed a decrease in performance when triples swapping are enabled. From further analysis of the dataset, we hypothesized that the inconsistent observations on both datasets mainly originate from the quality of seed alignment annotations. Specifically, when we connect all entities with seed alignments annotations to form an ``alignment graph'', we discovered several connected components of size over 1000 in such a graph on the E-PKG dataset, suggesting that over 1000 entities across the six KGs are annotated to be aligned to each other. This could either be due to incorrect and noisy annotations or potentially duplicate entities within monolingual KGs. A similar effect was not observed in DBP-5L. The effect of lower quality KGs and seed alignments was enlarged when triples generated from alignment-informed meta-path were introduced, generating a prohibitively large number of inferred triples, resulting in the inconsistent observations in table \ref{table:meta}. Compared to DBP-5L, which was generated from the sampling of the widely adopted and verified DBpedia \citep{auer_dbpedia_2007}, E-PKG was a recently constructed industrial E-commerce product KG dataset \citep{huang_multilingual_2022}, which represents the quality of real-life industrial dataset to some extent. Therefore, we have decided not to make any assumptions about the dataset's quality in our settings and report the main results without including triples generated from alignment-informed meta-paths in our training.

\begin{figure}
  \centering
  \includegraphics[width=0.4\textwidth]{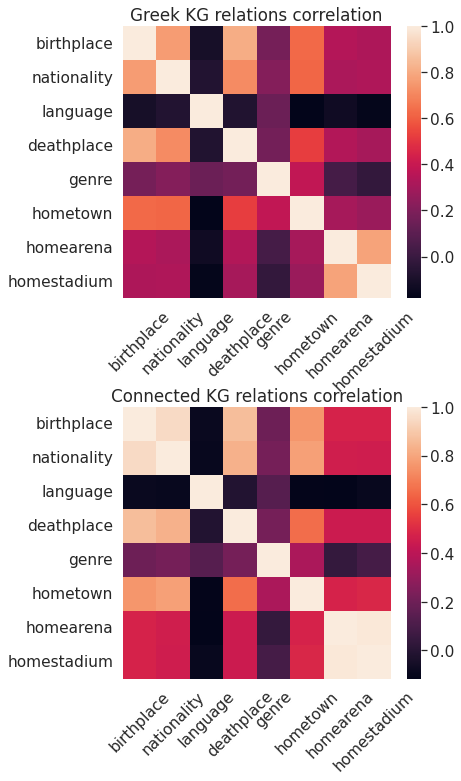}
  \caption{Relation correlation heatmap of Greek KG v.s. fused KG.}
  \label{figure:corr}
\end{figure}

\subsection{Knowledge transferred across KG}
    In this section, we discuss the knowledge being transferred across KGs. 
    Compared to the individual models trained on monolingual KGs, the model on fused KG implicitly captures the transferable triples via alignment information discussed previously while avoiding the error propagation of the potential noisy seed alignments to some extent (as shown in table \ref{table:meta}). Additionally, relations embeddings are shared across individual KGs within the connected model. As a result, relationships between the KG relations are also better captured. In figure \ref{figure:corr}, we visualize the correlations of relation embeddings learnt on Greek KG and the fused KG on the DBP-5L dataset. The correlations of learnt KG relations in the Greek KG differ from those discovered on the fused KG. Specifically, pairs of relatable connections like (``birthplace'', ``nationality''), (``birthplace'', ``deathplace''), as well as (``homearena'', ``homestadium''), all exhibit a higher correlation in the fused KG compared to the individual Greek KG. This implies that the connected model can capture more accurate dynamics between relations with more triples information and shared relation embeddings. At the same time, the low-resource Greek KG converged to a suboptimal region mainly due to the lack of information. The learnt knowledge is then transferred to the models on individual KGs via a knowledge distillation mechanism and further optimized during training. 

\section{Conclusions}
    In this paper, we proposed a novel method CKGC-CKD that focuses on the KGC task across multiple KGs. The proposed method uses an augmented CompGCN encoder for message propagation across different KGs via seed alignments in a fused KG. In addition, the proposed model employs additional mutual knowledge distillations between individual KGs and the fused KG to maximize knowledge transfer. CKGC-CKD outperforms the state-of-the-art models by a significant margin on the KGC task on multilingual datasets DBP-5L and E-PKG. We also demonstrate the performance gains provided by each component of the proposed method. One limitation of this work lies in the assumption of sufficient seed alignments between KGs. When seed alignments are more limited than current experiment settings, it is fairly obvious that our method has potentials to generate probalistic predictions of alignments aross KGs. As a result, we have planned future work to specifically focus on including more probabilistic/noisy alignments predictions iteratively while limiting the propagation of error in the iterative process.

\algnewcommand{\LineComment}[1]{\State \(\triangleright\) #1}
\begin{algorithm}
\caption{Pseudocode of the training process of CKGC-CKD.}\label{alg:cap}
\begin{algorithmic}
\LineComment{Stage 1: Train each model $M_i$ and $M_f$ with the Knowledge Representation loss.}
\For{$i \in 1..m+1$}
    \While{$M_i$ not converged}
         \State $L^i \gets T_i$
         \State $M_i \gets$ Update w.r.t $L^i$
    \EndWhile
\EndFor

\LineComment{Stage 2: Train each model $M_i$ and $M_f$ with the Knowledge Representation and the Knowledge Distillation loss.}
\While{not converged}
\State $batch_f \gets$ sample from triple set $T_f$
\State $L^f_T \gets$ calculate loss of $batch_f$ base on equation \ref{eq:kr}
    \For{$i \in 1..m$}
        \State $batch_i \gets$ sample from triple set $T_i$
        \State $L^i_T \gets$ calculate loss of $batch_i$ based on equation \ref{eq:kr}
        \State $L^i_D, L^f_D \gets$ calculate distillation losses between $M_i$ and $M_f$ on $batch_i$ based on equation \ref{eq:distill} with top-k sampling to select candidates space of distillation
        \State $L^i \gets L^i_T + \alpha L^i_D$
        \State $L^f \gets L^f_T + \alpha L^f_D$
        \State $M_i \gets$ Update w.r.t $L^i$
    \EndFor
    \State $M_f \gets$ Update w.r.t $L^f$
\EndWhile
\end{algorithmic}
\end{algorithm}

\bibliographystyle{ACM-Reference-Format}
\bibliography{enslp-22}

\end{document}